\pdfoutput=1

\documentclass[11pt]{article}

\usepackage[final]{acl}

\usepackage{times}
\usepackage{latexsym}
\usepackage{multirow}
\usepackage{float}
\usepackage{afterpage}
\usepackage{placeins}

\usepackage[T1]{fontenc}

\usepackage[utf8]{inputenc}

\usepackage{microtype}

\usepackage{inconsolata}

\usepackage{graphicx}
\usepackage{subfigure}

\usepackage[most]{tcolorbox}
\usepackage{xcolor}

\definecolor{lightgray}{gray}{0.9}
\definecolor{boxblue}{RGB}{220, 230, 241}
\definecolor{boxyellow}{RGB}{255, 244, 204}
\definecolor{lightgreen}{RGB}{220, 250, 220}
\definecolor{lightblue}{RGB}{210, 235, 250}

\title{Dataset of News Articles with Provenance Metadata \\ for Media Relevance Assessment}

\author{Tomas Peterka \\
  Gymnazium Jana Keplera \\
  \texttt{xpetto01@gjk.cz} \\ \And
  Matyas Bohacek \\
  Stanford University \\
  \texttt{maty@stanford.edu} \\}

\begin{document}
\maketitle
\begin{abstract}

Out-of-context and misattributed imagery is the leading form of media manipulation in today's misinformation and disinformation landscape. The existing methods attempting to detect this practice often only consider whether the semantics of the imagery corresponds to the text narrative, missing manipulation so long as the depicted objects or scenes somewhat correspond to the narrative at hand. To tackle this, we introduce \textit{News Media Provenance Dataset}, a dataset of news articles with provenance-tagged images. We formulate two tasks on this dataset, location of origin relevance (LOR) and date and time of origin relevance (DTOR), and present baseline results on six large language models (LLMs). We identify that, while the zero-shot performance on LOR is promising, the performance on DTOR hinders, leaving room for specialized architectures and future work.

\end{abstract}


\begin{figure*}[th]
    \centering
    \includegraphics[width=\linewidth]{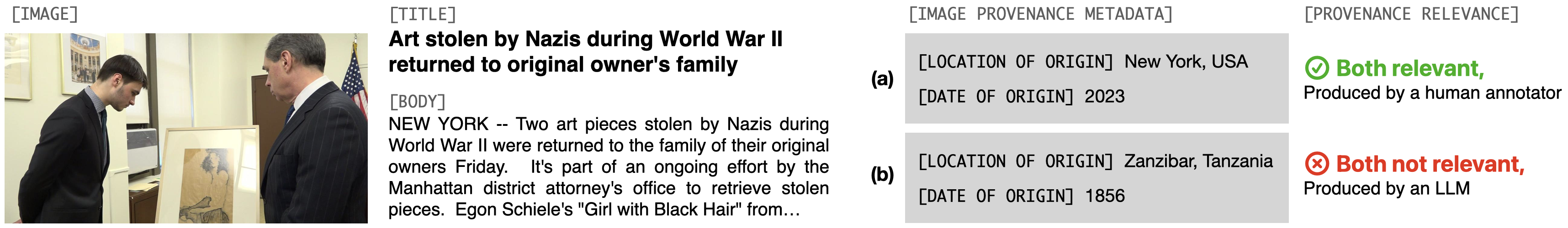}
    \caption{Representative example of a news article from the \textit{News Media Provenance Dataset} with a structured title, body, and image. This article appears in the dataset multiple times with alternative image provenance metadata, shown on the right. (a) One data point contains provenance metadata that was produced by a human annotator to match the relevance of the article. (b) Another data point contains provenance metadata that was randomly produced by an LLM to not match the relevance of the article. The article is sourced from CBS News.}
    \label{fig:example-data-point}
\end{figure*}

\section{Introduction}

Over the last few years, the use of manipulated imagery for disinformation and misinformation has grown steadily~\cite{dufour2024ammeba, shen2021photograph, weikmann2023visual, wang2024harmfully}. Many believe this is largely due to the abundance of AI-powered tools that allow users to edit or generate media from scratch, including images (text-to-image~\cite{baldridge2024imagen, bie2024renaissance, ramesh2021zero}, in-painting~\cite{liu2023image, lee2021restore}), audio (text-to-speech~\cite{eskimez2024e2, chen2024vall, lajszczak2024base}, voice cloning~\cite{qin2023openvoice, luong2020nautilus}), and video (deepfakes~\cite{pei2024deepfake, stanishevskii2024implicitdeepfake, croitoru2024deepfake}, text-to-video~\cite{singer2022make, zhang2025flashvideo}). These tools have not only become easily accessible online but also increasingly intuitive to use, often requiring only textual descriptions~\cite{rombach2022high}. Consequently, a large body of work has emerged focusing on the detection of AI-manipulated or AI-generated content~\cite{nguyen2022deep,farid2022creating}.

However, despite the proliferation of AI tools, a simpler form of image-based manipulation remains prevalent in misinformation and disinformation~\cite{garimella2020images}: the use of out-of-context or misattributed imagery to frame events in misleading ways~\cite{fazio2020out}. For example, in April 2020, images of body bags from Ecuador were falsely presented as deceased COVID-19 patients in New York hospitals~\cite{newslit}, sparking confusion and controversy online. Studies indicate that this type of manipulation appears in over $40\%$ of online misinformation containing images, whereas AI-generated media is used in approximately $30\%$~\cite{dufour2024ammeba}.

Despite this, the literature has not responded to the threat of out-of-context and misattributed imagery with the same urgency as AI-manipulated and AI-generated content. As a result, there is a scarcity of specialized resources---methods, tools, datasets, and benchmarks---for studying this phenomenon from the perspective of natural language processing (NLP). Some existing work evaluates whether an image is relevant to the article in which it appears, it primarily considers whether the depicted object or scene aligns with the textual narrative~\cite{aneja2021cosmos}. While this analyzes one aspect of media-based manipulation, it misses cases where the imagery and text appear semantically consistent but were captured at times or places that may be irrelevant or outright deceptive.

\citet{peterka2025large}, therefore, suggest a new formulation of this task. Rather than asking \textit{"Is this image relevant to the news story?"}, they instead ask \textit{"Was this image captured at a time and place that is relevant to the news story?"}. To this end, they hypothesize that provenance metadata---a record of a file's existence from its creation through edits to distribution---could help answer this question. Hence, they conduct some exploratory experiments with large language models (LLMs) to analyze the metadata of images used in news articles. However, they identify two major limitations: (1) the absence of a benchmark dataset for this task and (2) the early-stage adoption of provenance metadata among news outlets, restricting robust evaluation.

In response, we introduce a dataset of news articles with provenance-tagged images and annotations regarding their relevance to the article. Since the news outlets from which the articles were sourced do not yet incorporate provenance metadata (consistent with the limitation identified above), we simulate it. Specifically, we gather annotations for relevant locations and dates and embed them into the images using C2PA~\cite{rosenthol2022c2pa}, a widely used provenance metadata library. We then use an LLM to generate alternative, non-relevant dates and locations, constructing a balanced dataset containing relevant, partially relevant, and irrelevant images based on provenance.

While provenance metadata is not limited to images, our dataset and evaluations focus exclusively on news articles with images. Other modalities, such as video or audio, are not included, since the modality of the file from which provenance metadata is extracted does not affect the included information. 

The primary contributions of this paper can be summarized as follows:

\begin{itemize}
    \item We introduce the first news dataset with provenance metadata-equipped images, \textit{News Media Provenance Dataset}, and open-source\footnote{\url{https://huggingface.co/datasets/matybohacek/News-Media-Provenance-Dataset}} it for research use.
    \item We propose two provenance-based tasks with applications beyond news and authenticity analysis: (1) location of origin relevance (LOR) assessment and (2) date and time of origin relevance (DTOR) assessment.
    \item We report baseline results of six LLMs and detail a qualitative assessment of their shortcomings, with fully open-sourced\footnote{\url{https://news-provenance.github.io}} experimental scripts and prediction data.
\end{itemize}

\section{Related Work}

This section reviews existing NLP literature connected to image and video relevance assessment in news articles. First, we provide an overview of the broader area of study, which involves news articles in NLP. We then proceed specifically to existing work on image and video relevance and data provenance.

\subsection{News-Specific Tasks and Datasets}

News articles have become a productive subject of study in the NLP community, as they are largely abundant, reflective of current discourse, and invite many direct applications of NLP technology. We categorize some of the most prominent works in this domain by the nature of their task.

\subsubsection{Text Classification}

There is a robust body of work pertaining to news article classification---spanning topic categories, sentiment analysis, political tendencies, and more. Prominent datasets for this task category include AG News~\cite{gulli2005anatomy} with $120,000$ articles, 20 Newsgroups~\cite{lang1995newsweeder} with $18,000$ articles, Reuters-21578 with $21,000$ articles focused on finance, News Category Dataset~\cite{misra2022news} with $210,000$ articles from HuffPost, Multilabeled News Dataset (MN-DS)~\cite{petukhova2023mn} with $10,000$ articles across $215$ news sources, and KINNEWS/KIRNEWS~\cite{niyongabo2020kinnews} with $3,000$ tailored for low-resource African languages.

\subsubsection{Summarization}

Another prominent task involving news articles is summarization, attempting to reduce the full article body into a concise abstract while preserving the core information value. Prominent datasets for this task category include CNN/DailyMail~\cite{hermann2015teaching} with $287,000$ article-highlight pairs, NEWSROOM~\cite{grusky2018newsroom} with $1.3\text{M}$ articles across $38$ news sources, CCSUM~\cite{jiang2024ccsum}, with $1.3\text{M}$ articles, and SumeCzech~\cite{straka2018sumeczech} with $1\text{M}$ Czech articles.

\subsubsection{Disinformation Detection}

In the last few years, disinformation detection (also referred to as fake news detection) has emerged as a productive area of study in the literature. The framing of the problem varies both on the side of category definitions (what constitutes disinformation and how to categorize its severity) and on the side of modeling (approaches range from classification to feature detection to question answering).

Prominent datasets for this task category include the LIAR benchmark~\cite{wang2017liar} with over $12,000$ articles, the Verifee dataset~\cite{bohacek2023czech} with over $10,000$ articles spanning $60$ news sources, NELA-GT~\cite{gruppi2021nela} with $713,000$ articles, and FNC-1~\cite{slovikovskaya2019transfer} with $49,972$ articles.

\subsection{Image and Video Relevance in News}

Next, we review previous work specifically targeting the relevance of imagery in news articles.

\citet{cheema2023understanding} were among the first to explore computational approaches to modeling this relationship between imagery and news articles with modern NLP techniques. Their work, however, primarily set out to review the landscape of existing methods at the time and assess the overall feasibility of future methods in the area; the paper is, hence, primarily descriptive and does not present a specific dataset or architecture.

\citet{tonglet2024image} materialized many of the dynamics described by \citet{cheema2023understanding} by using a VLM to ask questions about the thumbnail image, deriving its relevance to the rest of the article. However, these inferences are based purely on LLM predictions, and so imagery presenting semantically relevant events may pass the test even when taken at an irrelevant time or place.

Later, \citet{yoon2024understanding} proposed CFT-CLIP, a framework evaluating the relevance of thumbnail images with respect to the remaining text based on multimodal embeddings. To that end, they also introduced a curated dataset called NewsTT, which contains $1,000$ annotated news image-text pairs with relevance labels. This method, however, only reflects the relevance of an image based on its semantic distance from the text, disregarding when and where the image was taken.

Finally, \citet{aneja2021cosmos} introduced the COSMOS dataset for out-of-context thumbnail image detection, enriched by captions with named entity labels. The authors also proposed a self-supervised architecture tailored to this task. While this dataset is concerned with the relevance of media in news articles, as are we, it is, yet again, based on semantical consistency or divergence between the semantics of the image and its caption.

\begin{figure*}[th]
    \centering
    \includegraphics[width=\linewidth]{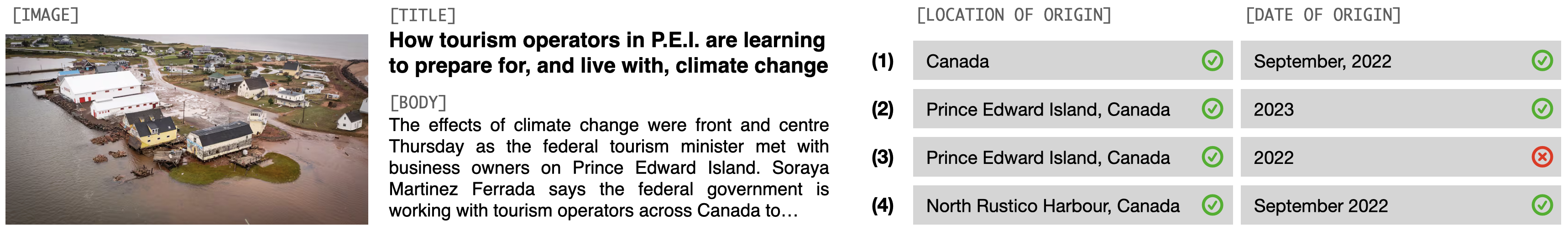}
    \\
    \vspace{0.3cm}
    \includegraphics[width=\linewidth]{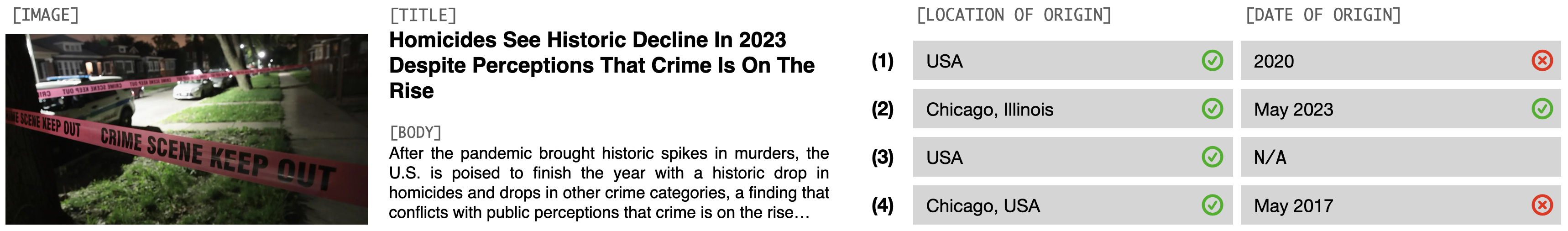}
    \caption{Examples of images from the \textit{News Media Provenance Dataset} used to evaluate annotator reliability. All four annotators provided the location and date of origin for each image, with their accuracy indicated on the right. The article at the top is sourced from CBC and the article at the bottom is sourced from Forbes.}
    \label{fig:annotator-reliability}
\end{figure*}

\subsection{Data Provenance}

Moving beyond semantics inferred from pixels, data provenance can offer information about the origin, evolution, and ownership of a piece of data. While specific implementations of data provenance metadata vary in the covered scope of information, underlying transaction mechanisms, and security guarantees, most existing frameworks include the location and date/time of origin of the data. The framework that has recognized the most adoption by social media platforms, newspapers, and tech companies to date, as compared to alternatives, is C2PA~\cite{rosenthol2022c2pa}, which we adopt in this paper.

While C2PA offers advantages such as guarantees of cryptographic security and unstrippable metadata technology, it has multiple limitations~\cite{longpre2024data, c2pa}. The primary limitation hindering adoption is that most digital content today lacks C2PA provenance metadata. As a result, any analysis dependent on C2PA remains infeasible for the majority of online content. While this may be prohibitive for existing consumer-facing applications, the adoption of C2PA and similar frameworks has been increasing, and so we can expect that, in the future, such analysis will be feasible.

Given the cryptographic guarantees for establishing the trace of an image or a video, which provenance metadata enables, it seems highly desirable for relevance assessment of imagery in news articles. To the best of our knowledge, no datasets or resources currently exist for evaluating provenance in news articles.

\section{News Media Provenance Dataset}

This section presents the \textit{News Media Provenance Dataset}, comprising $637$ news articles with simulated image provenance metadata, which is labeled either as \textit{relevant} or \textit{not relevant}. The provenance is inserted into the images using the C2PA library~\cite{rosenthol2022c2pa} by us: the relevant information is provided by annotators and the not relevant is generated using an LLM. Two example data points are shown in Figure~\ref{fig:example-data-point}.


\subsection{Dataset Construction}

This section reviews the dataset construction including data sourcing, filtering, and annotations management. The code used for these tasks is fully open-sourced\footnote{\url{https://news-provenance.github.io}}. Any modifications to default library behavior mentioned below are further expanded upon in the documentation of the code release.

\subsubsection{Data collection}

A list of news article URLs was obtained from the the Webz.io News Dataset Repository~\cite{webhose} in November 2024. Newsarticle4k~\cite{ou2013newspaper3k, newspaper4k} with custom extensions was then used to loop over these article URLs (in randomized order), extracting structured information from the website: the title, body, main image, and its caption. This loop terminated once $200$ news articles were successfully scraped.

\subsubsection{Annotation Procedure}

Four annotators were recruited through Prolific to simulate relevant image provenance metadata for the $200$ scraped articles. Out of these annotators, two were male and two were female, ranging in age from $23$ to $31$. All were based in the United States and we paid them $12$ USD per hour.

Each annotator was assigned $55$ articles. The first five were shared across all annotators for annotator reliability evaluation; the remaining $50$ were unique to the annotator.

The annotations were facilitated through the Argilla\footnote{\url{https://argilla.io}} tool. Representative screenshots of the tool are presented in Figures~\ref{fig:ann-tool-0}-\ref{fig:ann-tool-2} (Appendix~\ref{app:annotation-tool-screenshots}). It took the annotators, on average, $60$ minutes to annotate all the assigned articles. This excludes the time spent familiarizing themselves with the annotation instructions and set up the interface.


\begin{figure*}[t]
    \centering
    \subfigure{
        \includegraphics[width=0.31\textwidth]{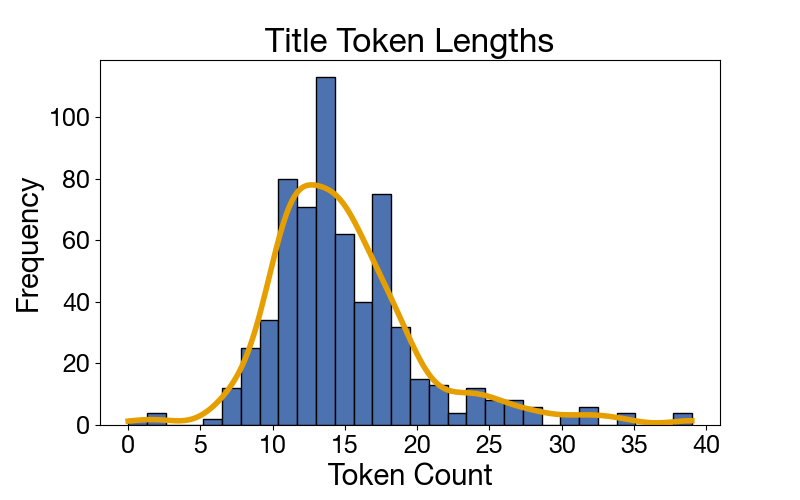}
    }
    \subfigure{
        \includegraphics[width=0.31\textwidth]{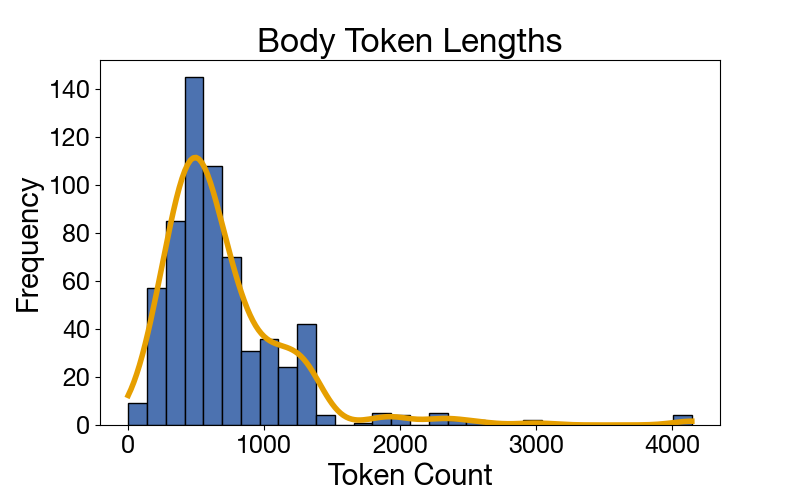}
    }
    \subfigure{
        \includegraphics[width=0.31\textwidth]{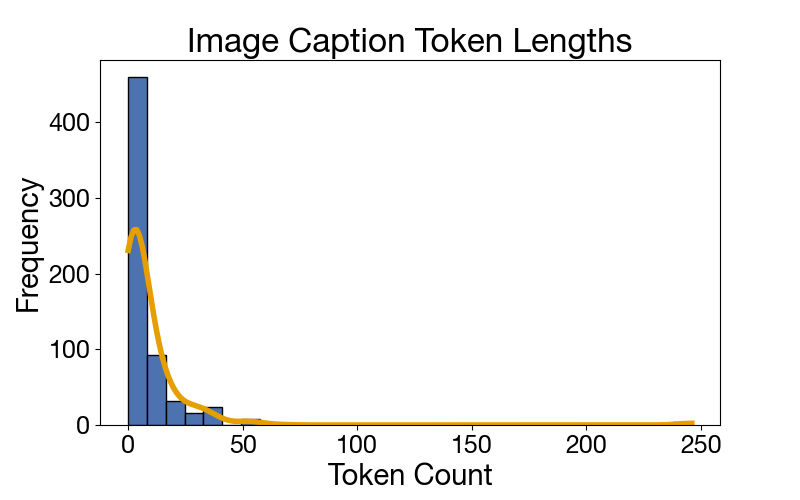}
    }
    \caption{Distribution of the title, body, and image caption token length in the \textit{News Media Provenance Dataset}. A fitted Kernel Density Estimation (KDE) is shown in orange. Outliers were manually reviewed to prevent scraping issues.}
    \label{fig:token-length-stats}
\end{figure*}

\subsubsection{Annotation Reliability}

The annotator reliability was evaluated on the first five articles which were assigned to all annotators. The annotators provided the correct location of origin in $80\%$ of the cases and the correct date of origin in $56\%$ of the cases.\footnote{This discounts cases in which the user deemed the attribute as ambiguous and responded with N/A. We allowed a $\pm1$ buffer for the date of origin units.}

Examples of these articles alongside annotator responses are shown in Figure~\ref{fig:annotator-reliability}. The article at the top had an solid annotator performance; the article at the bottom had a somewhat poor annotator performance on the date of origin. Note that the level of detail of both the provided location and date differ; as long as all components match, the response is deemed as correct. 






\subsection{Alternative Provenance Generation}

While the annotations served to simulate provenance metadata where both the location and date and time of origin are relevant to the articles, ChatGPT-4o~\cite{hurst2024gpt} was used to simulate additional provenance metadata that were not relevant to the article. With the prompt presented in Appendix~\ref{app:prompts}, the model was asked to generate three additional data points:\footnote{If either annotation was N/A, then the generation of respective matches (that are not relevant to the article) was skipped.} two data points where one of the provenance metadata fields is not relevant but the other is kept intact, and one data point where both provenance metadata fields are not relevant.

\subsection{Dataset Statistics}

\begin{figure}
    \centering
    \includegraphics[width=\linewidth]{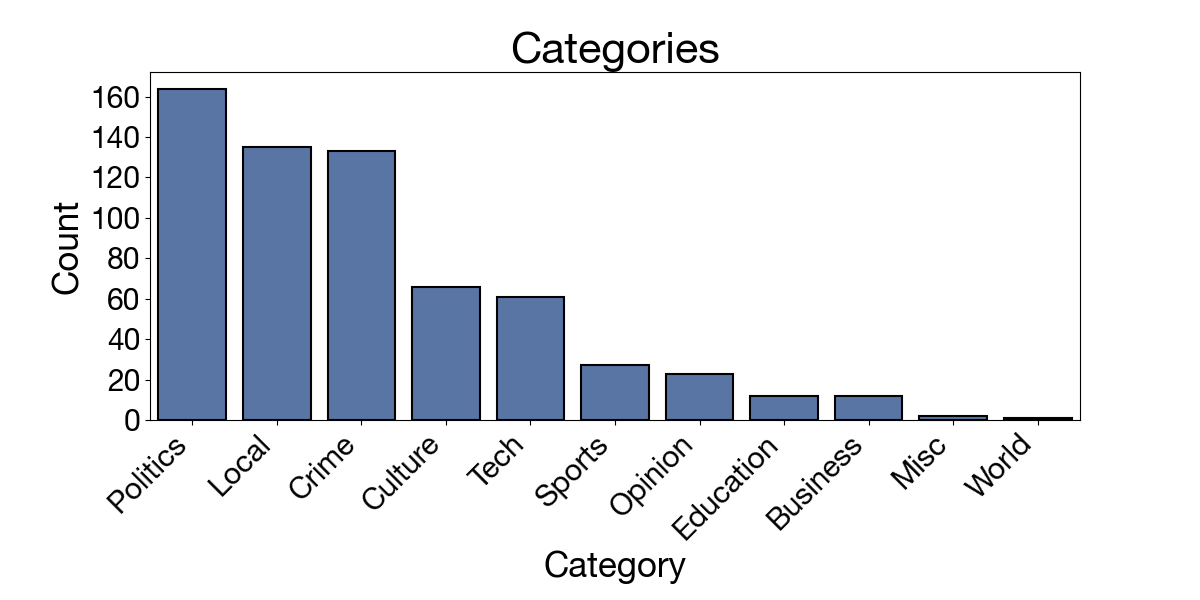}
    \caption{Distribution of categories in the \textit{News Media Provenance Dataset}. A single news article (data point) is represented only once by its primary category.}
    \label{fig:category-distribution}
\end{figure}

\begin{figure}
    \centering
    \includegraphics[width=\linewidth]{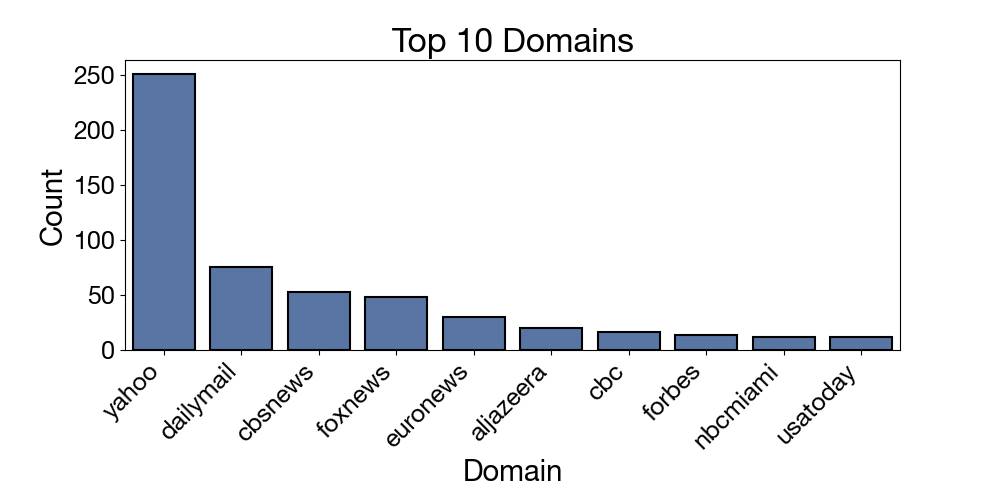}
    \caption{Distribution of source domains in the \textit{News Media Provenance Dataset}, showing the top $10$ domains.}
    \label{fig:domain-distribution}
\end{figure}

In total, the dataset contains $637$ news articles. Their length statistics are shown in Figure~\ref{fig:token-length-stats}. The average length of the headline, body, and image caption, calculated with NLTK~\cite{bird2006nltk}, are $15$, $705$, and $9$ tokens, respectively.

The top-$10$ domains by absolute article count are \texttt{yahoo}, \texttt{dailymail}, \texttt{cbsnews}, \texttt{foxnews}, \texttt{euronews}, \texttt{aljazeera}, \texttt{cbc}, \texttt{forbes}, \texttt{nbcmiami}, and \texttt{usatoday}, as shown in Figure~\ref{fig:token-length-stats}. There appears to be an imbalance of \texttt{yahoo}-domain articles. We investigated this, but found that it is because \texttt{yahoo} republishes news articles from other domains, and that the actual source distribution among these articles is diverse. We, hence, did not pursue any balancing remedies.

The category statistics, as predicted by a one-shot text classification model~\cite{lewis2019bart}, are shown in Figure~\ref{fig:category-distribution}. The majority of articles in the dataset fall within the category of Politics, Local, and Crime news.

\begin{table*}[t]
\centering
\begin{tabular}{l|cc|ccc}
\multirow{2}{*}{\textbf{Model}} & \multicolumn{2}{c|}{\textbf{Feature-level}} & \multicolumn{3}{c}{\textbf{Article-level}} \\
 & LOR & DTOR & 2 corr & 1 corr & 0 corr \\
\hline
 ChatGPT-4o & \textbf{0.81} & 0.57 & \textbf{0.45} & 0.47 & 0.08 \\ 
 DeepSeek V3 & 0.69 & 0.56 & 0.36 & 0.54 & 0.10 \\ 
 Gemma 2 27B Instruct & 0.77 & \textbf{0.58} & 0.41 & 0.53 & 0.06 \\ 
 Llama 3.1 8B Instruct & 0.64 & 0.42 & 0.24 & 0.57 & 0.19 \\ 
 Mistral 7B Instruct v0.3 & 0.73 & 0.47 & 0.32 & 0.56 & 0.12 \\ 
 Phi 3.5 Vision Instruct & 0.64 & 0.48 & 0.30 & 0.53 & 0.17 \\
\end{tabular}
\caption{Accuracy of baseline LLMs on the newly proposed LOR and DTOR (feature-level) tasks using the \textit{News Media Provenance Dataset}. The article-level statistics indicate the proportion of articles where both LOR and DTOR predictions were correct (2 corr), one of the predictions was correct (1 corr), and no prediction was correct (0 corr).}
\label{tab:model_benchmarking}
\end{table*}



\subsection{Proposed Tasks}

We propose two tasks on the dataset: \textit{Location of Origin Relevance (LOR)} assessment and \textit{Date and Time of Origin Relevance (DTOR)} assessment. Note that, while the image was presented to the annotators, these tasks do not assume access to the image. The purpose of these tasks is not to assess whether the semantics of the image (inferred from the pixel space) are relevant to the topic, but rather whether the circumstances, in which the image was captured, are relevant to the presented article.

\subsubsection{Location of Origin Relevance (LOR)}

The LOR task comprises the following: given the main image's location of origin found in the provenance metadata, determine whether the image is relevant to the article (represented as title and body).

\subsubsection{Date and Time of Origin Relevance (DTOR)}

The DTOR task comprises the following: given the main image's date and time of origin found in the provenance metadata, determine whether the image is relevant to the article (represented as title and body).

\section{Baseline Models}

We evaluate the following off-the-shelf LLMs to establish baseline results: ChatGPT-4o~\cite{hurst2024gpt}, DeepSeek V3~\cite{liu2024deepseek}, Gemma 2 27B Instruct~\cite{team2024gemma}, Llama 3.1 8B Instruct~\cite{dubey2024llama}, Mistral 7B Instruct~\cite{jiang2023mistral}, and Phi 3.5 Vision Instruct~\cite{abdin2024phi}. These are some of the most prominent models in the community, chosen based on their popularity on Hugging Face Transformers~\cite{wolf2020transformers} and overall benchmark performance at the time of writing.

Note that the parameter size and training scope of these models vary, and one can, of course, expect the larger models to outperform the smaller ones. For example, it is reasonable to expect that ChatGPT-4o or Deepseek V3 will outperform the much smaller Llama 3.1 8B Instruct. The results of this analysis should serve as a baseline for future work investigating methods designed specifically for LOR and DTOR. 

The ChatGPT-4o inference was performed using OpenAI's API. The remaining models were implemented using the Hugging Face Transformers~\cite{wolf2020transformers} library. To preserve some comparability across models, all inference parameters were left at their default values, and thus mimicking off-the-shelf use. The full prompt is presented in Appendix~\ref{app:prompts}. 

The binary responses to LOR and DTOR were converted from text to corresponding boolean representations. Whenever the LLM returned a response that did not conform to the JSON format specified in the prompt, the inference was repeated. The inference code and prediction data is open-sourced\footnote{\url{https://news-provenance.github.io}} to maximize reproducibility.

\section{Evaluation}

This section presents both quantitative and qualitative results of the baseline models evaluated on the \textit{News Media Provenance Dataset}.

\subsection{Quantitative Evaluation}
\label{sec:results-quant}

Table~\ref{tab:model_benchmarking} presents the LOR and DTOR accuracy for all evaluated models. LOR performance ranges from $64\%$ to $81\%$, with the highest accuracy achieved by ChatGPT-4o. Close behind are Gemma 2 27B Instruct at $77\%$, Mistral 7B Instruct v0.3 at $73\%$, and DeepSeek V3 at $69\%$. Llama 3.1 8B Instruct and Phi 3.5 Vision Instruct both attain an accuracy of $64\%$.

DTOR performance ranges from $42\%$ to $58\%$, with the highest accuracy achieved by Gemma 2 27B Instruct. Compared to LOR, accuracies on this task are generally lower, indicating that determining the relevance of date and time of origin is more challenging than assessing location relevance. While the three larger models---ChatGPT-4o, DeepSeek V3, and Gemma 2 27B Instruct—performed comparatively on this task (with ChatGPT-4o and DeepSeek V3 achieving $57\%$ and $56\%$, respectively), the smaller models---Llama 3.1 8B Instruct, Mistral 7B Instruct v0.3, and Phi 3.5 Vision Instruct—scored at or below $50\%$. This suggests that their representations or architectures either fail to capture relevant signals for this task or even lead to counterproductive predictions.

Table~\ref{tab:model_benchmarking} also reports aggregate article-level performance for the evaluated models. The best-performing model in this regard is ChatGPT-4o, which correctly predicted both LOR and DTOR for $45\%$ of the articles, predicted only one correctly $47\%$ of the time, and failed on both $7\%$ of the time. Overall performance across models is relatively low, primarily due to poor DTOR performance, which significantly impacts the aggregate metrics.

\subsection{Qualitative Evaluation}
\label{sec:results-qual}

For ChatGPT-4o, the best-performing model in the above quantitative evaluation, we manually reviewed all article analyses produced during the evaluation. In two follow-up questions, we asked the model to support its reasoning. This approach aimed to better understand the model’s ability to reason about LOR and DTOR.

To illustrate the overall trend we observed across many articles, consider an article titled \textit{"N.Y. sushi restaurant owner out of coma after dine-and-dash attack over \$425 bill"}, published on NBC News (see Appendix~\ref{app:qual-res-article}). When presented to ChatGPT-4o for evaluation with relevant location metadata (Albany, USA) and date and time metadata (May, 2024), it predicted yes for LOR (which is correct) and no for DTOR (which is incorrect).

We asked the model to provide reasoning on the LOR analysis as follows:

\begin{tcolorbox}[colback=lightgray, colframe=black, title=LOR Follow-up Prompt]
Provide reasoning for your LOR assessment.
\end{tcolorbox}

The model responded with the following:

\begin{tcolorbox}[colback=lightblue, colframe=black, title=LOR Reasoning Response]
The image was taken in Albany, USA, which is directly relevant to the news story. The incident occurred in Albany, specifically outside Shogun Sushi, making the location pertinent to the events described.
\end{tcolorbox}

This suggests that the model directly compared the location described in the article to the provenance metadata, as intended. This example reflects a broader pattern we observed in the data: the model was usually able to extract relevant locations and compare them to the provenance metadata according to the instructions.

We followed with another question:

\begin{tcolorbox}[colback=lightgray, colframe=black, title=DTOR Follow-up Prompt]
Provide reasoning for your DTOR assessment.
\end{tcolorbox}

The model responded with the following:

\begin{tcolorbox}[colback=lightblue, colframe=black, title=DTOR Reasoning Response]
The image was taken in May 2024, which is a bit ambiguous in terms of relevance. While the attack occurred on May 29, the story also mentions events happening after that—Su Wen waking from a coma and being moved out of ICU occurred later in June 2024. Since it is unclear whether the image was taken before or after the attack, and the article focuses heavily on recovery updates after the incident, the timing is not conclusively relevant to the most significant developments in the article.
\end{tcolorbox}

This indicates that the model made additional assumptions about the task, while lacking the ability to distinguish between the article’s temporal perspective and that of the analysis.

This reflects a broader tendency we observed: the model was largely unable to reason soundly about dates and timelines, leading to many incorrect inferences. It also often conflated days of the week with months and years in a way that is not relevant to DTOR analysis.

\section{Discussion}

This section first discusses the significance of the results attained and then shifts to examining the broader implications of our dataset and the newly proposed tasks.

The quantitative results presented in Section~\ref{sec:results-quant} suggest that current LLMs can perform LOR assessment in a zero-shot setting and achieve strong baseline performance. This is not the case, however, for DTOR, where the evaluated models struggled to achieve accuracy above $50\%$. These findings are supported by our qualitative results, reported in Section~\ref{sec:results-qual}, which show that while the models can reason soundly about the location presented in the article, they struggle with reasoning about dates and timelines. This highlights a broader limitation of LLMs and underscores the need for further research into improving temporal reasoning capabilities.

In addition to challenges with representing time, we also observed that more recent news articles were often more difficult for the models to reason about. We hypothesize that this may stem from the nature of the models' training, as the most recent events are typically not included in their training datasets, making it harder for them to process or contextualize such information.

As expected, larger models outperformed smaller models in our evaluation. The performance of each model could likely be improved by optimizing its parameters and customizing the instruction prompts. We, however, chose to pursue minimal optimization to maintain a level of comparability necessary for measuing baseline results. The relatively low baseline performance nonetheless reinforces the need for developing new architectures tailored to the LOR and DTOR tasks.

We expect our dataset to play a critical role in this effort, as, to the best of our knowledge, there are no other datasets explicitly designed for the tasks of LOR and DTOR. Expanding the dataset to include non-Western news contexts and additional languages will also be essential to ensure inclusive support for underserved communities, who are often at greater risk of media manipulation.



\section{Limitations}

Despite the benefits of provenance metadata for assessing the relevance of media in news articles, some limitations remain. One major issue is that, even when an image or video presented alongside an article matches the scope and timeline of the story, the article can still be inaccurate or outright manipulative. We, therefore, see our method as just one tool that should be a part of a broader suite of techniques aimed at discerning problematic practices in news articles.

C2PA, the employed provenance metadata framework, also has some drawbacks. Older photos usually lack provenance data, limiting the use of our method on historical images. Moreover, there are articles in which the presence of time- and location-matched media is not necessarily an indicator of relevance. An example of this would be articles reporting on events without clearly bounded locations and/or time frames, such as natural disasters, which often span broad regions and extended periods. Additionally, certain media can be used for illustrative purposes, where strict provenance alignment is less critical to the integrity of the article (e.g., historical illustrations or generic portraits). In such cases, assessing metadata relevance requires a more flexible, nuanced approach. Future work could explore automatic methods for detecting when precise alignment is necessary. Furthermore, as C2PA is still a new technology, its adoption among media organizations is still limited. With many outlets pledging to join, however, its use is expected to grow.

\section{Ethical and Societal Implications}

The use of provenance metadata for assessing the relevance of media in news articles raises ethical concerns pertaining privacy. Embedding provenance metadata includes potentially sensitive information, such as location and device information, that could put journalists and activists reporting from unsafe regions at risk. Sharing any information that could reveal identity or location of individuals in such contexts may be undesirable and, we believe, should take priority over establishing trustworthy news channels. 

This also leads to a broader point, which we wish to highlight. Even though we gathered feedback on our approach from both practitioners and scholars of journalism, there may be additional implications for journalists and their readers. We, therefore, recommend that before this method (or its derivatives) are put in use at a news organization, they should be first extensively scrutinized by its staff to uncover any additional concerns.

Simultaneously, we remain optimistic that this method will introduce an effective tool to support individuals in an increasingly less credible and transparent information ecosystem. To that end, we believe our dataset will serve as a critical tool to improve and evaluate approaches to LOR and DTOR moving forward.

\section{Conclusion}

This paper defined the tasks of Location of Origin Relevance (LOR) and Date and Time of Origin Relevance (DTOR) for media (images and videos) presented alongside news articles, based on their provenance metadata. Since no suitable datasets existed for these tasks, we introduced the \textit{News Media Provenance Dataset}---a collection of news articles with provenance-tagged images---containing both human-annotated relevant metadata and irrelevant metadata generated by a large language model (LLM). We presented baseline zero-shot results for six prominent LLMs and found that, while out-of-the-box LOR performance is strong, DTOR performance remains limited, as models struggle to reason about time relevance and temporal relationships.




\bibliography{custom}

\begin{thebibliography}{58}
\providecommand{\natexlab}[1]{#1}

\bibitem[{Abdin et~al.(2024)Abdin, Aneja, Awadalla, Awadallah, Awan, Bach, Bahree, Bakhtiari, Bao, Behl et~al.}]{abdin2024phi}
Marah Abdin, Jyoti Aneja, Hany Awadalla, Ahmed Awadallah, Ammar~Ahmad Awan, Nguyen Bach, Amit Bahree, Arash Bakhtiari, Jianmin Bao, Harkirat Behl, et~al. 2024.
\newblock Phi-3 technical report: A highly capable language model locally on your phone.
\newblock \emph{arXiv preprint arXiv:2404.14219}.

\bibitem[{AndyTheFactory(2023)}]{newspaper4k}
AndyTheFactory. 2023.
\newblock \href {https://github.com/AndyTheFactory/newspaper4k} {{Newspaper4k}: {A}rticle scraping \& curation}.

\bibitem[{Aneja et~al.(2021)Aneja, Bregler, and Nie{\ss}ner}]{aneja2021cosmos}
Shivangi Aneja, Chris Bregler, and Matthias Nie{\ss}ner. 2021.
\newblock Cosmos: Catching out-of-context misinformation with self-supervised learning.
\newblock \emph{arXiv preprint arXiv:2101.06278}.

\bibitem[{Baldridge et~al.(2024)Baldridge, Bauer, Bhutani, Brichtova, Bunner, Castrejon, Chan, Chen, Dieleman, Du et~al.}]{baldridge2024imagen}
Jason Baldridge, Jakob Bauer, Mukul Bhutani, Nicole Brichtova, Andrew Bunner, Lluis Castrejon, Kelvin Chan, Yichang Chen, Sander Dieleman, Yuqing Du, et~al. 2024.
\newblock Imagen 3.
\newblock \emph{arXiv preprint arXiv:2408.07009}.

\bibitem[{Bie et~al.(2024)Bie, Yang, Zhou, Ghanem, Zhang, Yao, Wu, Holmes, Golnari, Clifton et~al.}]{bie2024renaissance}
Fengxiang Bie, Yibo Yang, Zhongzhu Zhou, Adam Ghanem, Minjia Zhang, Zhewei Yao, Xiaoxia Wu, Connor Holmes, Pareesa Golnari, David~A Clifton, et~al. 2024.
\newblock Renaissance: A survey into ai text-to-image generation in the era of large model.
\newblock \emph{IEEE Transactions on Pattern Analysis and Machine Intelligence}.

\bibitem[{Bird(2006)}]{bird2006nltk}
Steven Bird. 2006.
\newblock Nltk: the natural language toolkit.
\newblock In \emph{Proceedings of the COLING/ACL 2006 interactive presentation sessions}, pages 69--72.

\bibitem[{Bohacek et~al.(2023)Bohacek, Bravansky, Trhl{\'\i}k, and Moravec}]{bohacek2023czech}
Matyas Bohacek, Michal Bravansky, Filip Trhl{\'\i}k, and V{\'a}clav Moravec. 2023.
\newblock Czech-ing the news: Article trustworthiness dataset for czech.
\newblock In \emph{Proceedings of the 13th Workshop on Computational Approaches to Subjectivity, Sentiment, \& Social Media Analysis}, pages 96--109.

\bibitem[{Cheema et~al.(2023)Cheema, Hakimov, M{\"u}ller-Budack, Otto, Bateman, and Ewerth}]{cheema2023understanding}
Gullal~S Cheema, Sherzod Hakimov, Eric M{\"u}ller-Budack, Christian Otto, John~A Bateman, and Ralph Ewerth. 2023.
\newblock Understanding image-text relations and news values for multimodal news analysis.
\newblock \emph{Frontiers in artificial intelligence}, 6:1125533.

\bibitem[{Chen et~al.(2024)Chen, Liu, Zhou, Liu, Tan, Li, Zhao, Qian, and Wei}]{chen2024vall}
Sanyuan Chen, Shujie Liu, Long Zhou, Yanqing Liu, Xu~Tan, Jinyu Li, Sheng Zhao, Yao Qian, and Furu Wei. 2024.
\newblock {VALL-E 2}: {N}eural codec language models are human parity zero-shot text to speech synthesizers.
\newblock \emph{arXiv preprint arXiv:2406.05370}.

\bibitem[{{Coalition for Content Provenance and Authenticity (C2PA)}(2023)}]{c2pa}
{Coalition for Content Provenance and Authenticity (C2PA)}. 2023.
\newblock \href {https://c2pa.org/specifications/specifications/1.0/security/_attachments/Initial_Adoption_Assessment.pdf} {{Harms, Misuse, and Abuse: Initial Adoption Assessment}}.

\bibitem[{Croitoru et~al.(2024)Croitoru, Hiji, Hondru, Ristea, Irofti, Popescu, Rusu, Ionescu, Khan, and Shah}]{croitoru2024deepfake}
Florinel-Alin Croitoru, Andrei-Iulian Hiji, Vlad Hondru, Nicolae~Catalin Ristea, Paul Irofti, Marius Popescu, Cristian Rusu, Radu~Tudor Ionescu, Fahad~Shahbaz Khan, and Mubarak Shah. 2024.
\newblock Deepfake media generation and detection in the generative ai era: A survey and outlook.
\newblock \emph{arXiv preprint arXiv:2411.19537}.

\bibitem[{Dubey et~al.(2024)Dubey, Jauhri, Pandey, Kadian, Al-Dahle, Letman, Mathur, Schelten, Yang, Fan et~al.}]{dubey2024llama}
Abhimanyu Dubey, Abhinav Jauhri, Abhinav Pandey, Abhishek Kadian, Ahmad Al-Dahle, Aiesha Letman, Akhil Mathur, Alan Schelten, Amy Yang, Angela Fan, et~al. 2024.
\newblock The llama 3 herd of models.
\newblock \emph{arXiv preprint arXiv:2407.21783}.

\bibitem[{Dufour et~al.(2024)Dufour, Pathak, Samangouei, Hariri, Deshetti, Dudfield, Guess, Escayola, Tran, Babakar et~al.}]{dufour2024ammeba}
Nicholas Dufour, Arkanath Pathak, Pouya Samangouei, Nikki Hariri, Shashi Deshetti, Andrew Dudfield, Christopher Guess, Pablo~Hern{\'a}ndez Escayola, Bobby Tran, Mevan Babakar, et~al. 2024.
\newblock {AMMeBa}: {A} large-scale survey and dataset of media-based misinformation in-the-wild.
\newblock arXiv:2405.11697.

\bibitem[{Eskimez et~al.(2024)Eskimez, Wang, Thakker, Li, Tsai, Xiao, Yang, Zhu, Tang, Tan et~al.}]{eskimez2024e2}
Sefik~Emre Eskimez, Xiaofei Wang, Manthan Thakker, Canrun Li, Chung-Hsien Tsai, Zhen Xiao, Hemin Yang, Zirun Zhu, Min Tang, Xu~Tan, et~al. 2024.
\newblock {E2 TTS}: {E}mbarrassingly easy fully non-autoregressive zero-shot tts.
\newblock In \emph{2024 IEEE Spoken Language Technology Workshop (SLT)}, pages 682--689. IEEE.

\bibitem[{Farid(2022)}]{farid2022creating}
Hany Farid. 2022.
\newblock Creating, using, misusing, and detecting deep fakes.
\newblock \emph{Journal of Online Trust and Safety}, 1(4).

\bibitem[{Fazio(2020)}]{fazio2020out}
Lisa Fazio. 2020.
\newblock Out-of-context photos are a powerful low-tech form of misinformation.
\newblock \emph{The Conversation}, 14(1).

\bibitem[{Garimella and Eckles(2020)}]{garimella2020images}
Kiran Garimella and Dean Eckles. 2020.
\newblock Images and misinformation in political groups: Evidence from whatsapp in india.
\newblock \emph{arXiv preprint arXiv:2005.09784}.

\bibitem[{Gruppi et~al.(2021)Gruppi, Horne, and Adal{\i}}]{gruppi2021nela}
Maur{\'\i}cio Gruppi, Benjamin~D Horne, and Sibel Adal{\i}. 2021.
\newblock Nela-gt-2020: A large multi-labelled news dataset for the study of misinformation in news articles.
\newblock \emph{arXiv preprint arXiv:2102.04567}.

\bibitem[{Grusky et~al.(2018)Grusky, Naaman, and Artzi}]{grusky2018newsroom}
Max Grusky, Mor Naaman, and Yoav Artzi. 2018.
\newblock Newsroom: A dataset of 1.3 million summaries with diverse extractive strategies.
\newblock \emph{arXiv preprint arXiv:1804.11283}.

\bibitem[{Gulli(2005)}]{gulli2005anatomy}
Antonio Gulli. 2005.
\newblock The anatomy of a news search engine.
\newblock In \emph{Special interest tracks and posters of the 14th international conference on World Wide Web}, pages 880--881.

\bibitem[{Hermann et~al.(2015)Hermann, Kocisky, Grefenstette, Espeholt, Kay, Suleyman, and Blunsom}]{hermann2015teaching}
Karl~Moritz Hermann, Tomas Kocisky, Edward Grefenstette, Lasse Espeholt, Will Kay, Mustafa Suleyman, and Phil Blunsom. 2015.
\newblock Teaching machines to read and comprehend.
\newblock \emph{Advances in neural information processing systems}, 28.

\bibitem[{Hurst et~al.(2024)Hurst, Lerer, Goucher, Perelman, Ramesh, Clark, Ostrow, Welihinda, Hayes, Radford et~al.}]{hurst2024gpt}
Aaron Hurst, Adam Lerer, Adam~P Goucher, Adam Perelman, Aditya Ramesh, Aidan Clark, AJ~Ostrow, Akila Welihinda, Alan Hayes, Alec Radford, et~al. 2024.
\newblock Gpt-4o system card.
\newblock \emph{arXiv preprint arXiv:2410.21276}.

\bibitem[{Jiang et~al.(2023)Jiang, Sablayrolles, Mensch, Bamford, Chaplot, Casas, Bressand, Lengyel, Lample, Saulnier et~al.}]{jiang2023mistral}
Albert~Q Jiang, Alexandre Sablayrolles, Arthur Mensch, Chris Bamford, Devendra~Singh Chaplot, Diego de~las Casas, Florian Bressand, Gianna Lengyel, Guillaume Lample, Lucile Saulnier, et~al. 2023.
\newblock Mistral 7b.
\newblock \emph{arXiv preprint arXiv:2310.06825}.

\bibitem[{Jiang and Dreyer(2024)}]{jiang2024ccsum}
Xiang Jiang and Markus Dreyer. 2024.
\newblock Ccsum: A large-scale and high-quality dataset for abstractive news summarization.
\newblock In \emph{Proceedings of the 2024 Conference of the North American Chapter of the Association for Computational Linguistics: Human Language Technologies (Volume 1: Long Papers)}, pages 7299--7329.

\bibitem[{{\L}ajszczak et~al.(2024){\L}ajszczak, C{\'a}mbara, Li, Beyhan, Van~Korlaar, Yang, Joly, Mart{\'\i}n-Cortinas, Abbas, Michalski et~al.}]{lajszczak2024base}
Mateusz {\L}ajszczak, Guillermo C{\'a}mbara, Yang Li, Fatih Beyhan, Arent Van~Korlaar, Fan Yang, Arnaud Joly, {\'A}lvaro Mart{\'\i}n-Cortinas, Ammar Abbas, Adam Michalski, et~al. 2024.
\newblock {BASE TTS}: {L}essons from building a billion-parameter text-to-speech model on 100k hours of data.
\newblock \emph{arXiv preprint arXiv:2402.08093}.

\bibitem[{Lang(1995)}]{lang1995newsweeder}
Ken Lang. 1995.
\newblock Newsweeder: Learning to filter netnews.
\newblock In \emph{Machine learning proceedings 1995}, pages 331--339. Elsevier.

\bibitem[{Lee et~al.(2021)Lee, Kim, Kim, and Kim}]{lee2021restore}
Eunhye Lee, Jeongmu Kim, Jisu Kim, and Tae~Hyun Kim. 2021.
\newblock Restore from restored: Single-image inpainting.
\newblock \emph{arXiv preprint arXiv:2102.08078}.

\bibitem[{Lewis et~al.(2019)Lewis, Liu, Goyal, Ghazvininejad, Mohamed, Levy, Stoyanov, and Zettlemoyer}]{lewis2019bart}
Mike Lewis, Yinhan Liu, Naman Goyal, Marjan Ghazvininejad, Abdelrahman Mohamed, Omer Levy, Ves Stoyanov, and Luke Zettlemoyer. 2019.
\newblock {BART}: {D}enoising sequence-to-sequence pre-training for natural language generation, translation, and comprehension.
\newblock \emph{arXiv preprint arXiv:1910.13461}.

\bibitem[{Liu et~al.(2024)Liu, Feng, Xue, Wang, Wu, Lu, Zhao, Deng, Zhang, Ruan et~al.}]{liu2024deepseek}
Aixin Liu, Bei Feng, Bing Xue, Bingxuan Wang, Bochao Wu, Chengda Lu, Chenggang Zhao, Chengqi Deng, Chenyu Zhang, Chong Ruan, et~al. 2024.
\newblock Deepseek-v3 technical report.
\newblock \emph{arXiv preprint arXiv:2412.19437}.

\bibitem[{Liu et~al.(2023)Liu, Niepert, and Broeck}]{liu2023image}
Anji Liu, Mathias Niepert, and Guy Van~den Broeck. 2023.
\newblock Image inpainting via tractable steering of diffusion models.
\newblock \emph{arXiv preprint arXiv:2401.03349}.

\bibitem[{Longpre et~al.(2024)Longpre, Mahari, Obeng-Marnu, Brannon, South, Kabbara, and Pentland}]{longpre2024data}
Shayne Longpre, Robert Mahari, Naana Obeng-Marnu, William Brannon, Tobin South, Jad Kabbara, and Sandy Pentland. 2024.
\newblock Data authenticity, consent, and provenance for ai are all broken: What will it take to fix them?

\bibitem[{Luong and Yamagishi(2020)}]{luong2020nautilus}
Hieu-Thi Luong and Junichi Yamagishi. 2020.
\newblock Nautilus: a versatile voice cloning system.
\newblock \emph{IEEE/ACM Transactions on Audio, Speech, and Language Processing}, 28:2967--2981.

\bibitem[{Misra(2022)}]{misra2022news}
Rishabh Misra. 2022.
\newblock News category dataset.
\newblock \emph{arXiv preprint arXiv:2209.11429}.

\bibitem[{{News Literacy Project}(2025)}]{newslit}
{News Literacy Project}. 2025.
\newblock \href {https://newslit.org/tips-tools/covid-19-video-out-of-context/} {Covid-19 video taken out of context}.
\newblock Accessed: 2025-02-23.

\bibitem[{Nguyen et~al.(2022)Nguyen, Nguyen, Nguyen, Nguyen, Huynh-The, Nahavandi, Nguyen, Pham, and Nguyen}]{nguyen2022deep}
Thanh~Thi Nguyen, Quoc Viet~Hung Nguyen, Dung~Tien Nguyen, Duc~Thanh Nguyen, Thien Huynh-The, Saeid Nahavandi, Thanh~Tam Nguyen, Quoc-Viet Pham, and Cuong~M Nguyen. 2022.
\newblock Deep learning for deepfakes creation and detection: A survey.
\newblock \emph{Computer Vision and Image Understanding}, 223:103525.

\bibitem[{Niyongabo et~al.(2020)Niyongabo, Qu, Kreutzer, and Huang}]{niyongabo2020kinnews}
Rubungo~Andre Niyongabo, Hong Qu, Julia Kreutzer, and Li~Huang. 2020.
\newblock Kinnews and kirnews: Benchmarking cross-lingual text classification for kinyarwanda and kirundi.
\newblock \emph{arXiv preprint arXiv:2010.12174}.

\bibitem[{Ou-Yang(2013)}]{ou2013newspaper3k}
Lucas Ou-Yang. 2013.
\newblock {Newspaper3k}: {A}rticle scraping \& curation.
\newblock \emph{Newspaper3k: Article Scraping \& Curation-Newspaper 0.0. 2 Documentation}.

\bibitem[{Pei et~al.(2024)Pei, Zhang, Hu, Zhang, Wang, Wu, Zhai, Yang, Shen, and Tao}]{pei2024deepfake}
Gan Pei, Jiangning Zhang, Menghan Hu, Zhenyu Zhang, Chengjie Wang, Yunsheng Wu, Guangtao Zhai, Jian Yang, Chunhua Shen, and Dacheng Tao. 2024.
\newblock Deepfake generation and detection: A benchmark and survey.
\newblock \emph{arXiv preprint arXiv:2403.17881}.

\bibitem[{Peterka and Bohacek(2025)}]{peterka2025large}
Tomas Peterka and Matyas Bohacek. 2025.
\newblock Large language models and provenance metadata for determining the relevance of images and videos in news stories.
\newblock \emph{arXiv preprint arXiv:2502.09689}.

\bibitem[{Petukhova and Fachada(2023)}]{petukhova2023mn}
Alina Petukhova and Nuno Fachada. 2023.
\newblock Mn-ds: A multilabeled news dataset for news articles hierarchical classification.
\newblock \emph{Data}, 8(5):74.

\bibitem[{Qin et~al.(2023)Qin, Zhao, Yu, and Sun}]{qin2023openvoice}
Zengyi Qin, Wenliang Zhao, Xumin Yu, and Xin Sun. 2023.
\newblock Openvoice: Versatile instant voice cloning.
\newblock \emph{arXiv preprint arXiv:2312.01479}.

\bibitem[{Ramesh et~al.(2021)Ramesh, Pavlov, Goh, Gray, Voss, Radford, Chen, and Sutskever}]{ramesh2021zero}
Aditya Ramesh, Mikhail Pavlov, Gabriel Goh, Scott Gray, Chelsea Voss, Alec Radford, Mark Chen, and Ilya Sutskever. 2021.
\newblock Zero-shot text-to-image generation.
\newblock In \emph{International conference on machine learning}, pages 8821--8831. Pmlr.

\bibitem[{Rombach et~al.(2022)Rombach, Blattmann, Lorenz, Esser, and Ommer}]{rombach2022high}
Robin Rombach, Andreas Blattmann, Dominik Lorenz, Patrick Esser, and Bj{\"o}rn Ommer. 2022.
\newblock High-resolution image synthesis with latent diffusion models.
\newblock In \emph{Proceedings of the IEEE/CVF conference on computer vision and pattern recognition}, pages 10684--10695.

\bibitem[{Rosenthol(2022)}]{rosenthol2022c2pa}
Leonard Rosenthol. 2022.
\newblock C2pa: the world’s first industry standard for content provenance (conference presentation).
\newblock In \emph{Applications of Digital Image Processing XLV}, volume 12226, page 122260P. SPIE.

\bibitem[{Shen et~al.(2021)Shen, Kasra, and O'Brien}]{shen2021photograph}
Cuihua Shen, Mona Kasra, and James O'Brien. 2021.
\newblock This photograph has been altered: Testing the effectiveness of image forensic labeling on news image credibility.
\newblock \emph{arXiv preprint arXiv:2101.07951}.

\bibitem[{Singer et~al.(2022)Singer, Polyak, Hayes, Yin, An, Zhang, Hu, Yang, Ashual, Gafni et~al.}]{singer2022make}
Uriel Singer, Adam Polyak, Thomas Hayes, Xi~Yin, Jie An, Songyang Zhang, Qiyuan Hu, Harry Yang, Oron Ashual, Oran Gafni, et~al. 2022.
\newblock {Make-A-Video}: {T}ext-to-video generation without text-video data.
\newblock \emph{arXiv preprint arXiv:2209.14792}.

\bibitem[{Slovikovskaya(2019)}]{slovikovskaya2019transfer}
Valeriya Slovikovskaya. 2019.
\newblock Transfer learning from transformers to fake news challenge stance detection (fnc-1) task.
\newblock \emph{arXiv preprint arXiv:1910.14353}.

\bibitem[{Stanishevskii et~al.(2024)Stanishevskii, Steczkiewicz, Szczepanik, Tadeja, Tabor, and Spurek}]{stanishevskii2024implicitdeepfake}
Georgii Stanishevskii, Jakub Steczkiewicz, Tomasz Szczepanik, S{\l}awomir Tadeja, Jacek Tabor, and Przemys{\l}aw Spurek. 2024.
\newblock Implicitdeepfake: Plausible face-swapping through implicit deepfake generation using nerf and gaussian splatting.
\newblock \emph{arXiv e-prints}, pages arXiv--2402.

\bibitem[{Straka et~al.(2018)Straka, Mediankin, Kocmi, {\v{Z}}abokrtsk{\`y}, Hude{\v{c}}ek, and Hajic}]{straka2018sumeczech}
Milan Straka, Nikita Mediankin, Tom Kocmi, Zden{\v{e}}k {\v{Z}}abokrtsk{\`y}, Vojt{\v{e}}ch Hude{\v{c}}ek, and Jan Hajic. 2018.
\newblock Sumeczech: Large czech news-based summarization dataset.
\newblock In \emph{Proceedings of the Eleventh International Conference on Language Resources and Evaluation (LREC 2018)}.

\bibitem[{Team et~al.(2024)Team, Riviere, Pathak, Sessa, Hardin, Bhupatiraju, Hussenot, Mesnard, Shahriari, Ram{\'e} et~al.}]{team2024gemma}
Gemma Team, Morgane Riviere, Shreya Pathak, Pier~Giuseppe Sessa, Cassidy Hardin, Surya Bhupatiraju, L{\'e}onard Hussenot, Thomas Mesnard, Bobak Shahriari, Alexandre Ram{\'e}, et~al. 2024.
\newblock Gemma 2: Improving open language models at a practical size.
\newblock \emph{arXiv preprint arXiv:2408.00118}.

\bibitem[{Tonglet et~al.(2024)Tonglet, Moens, and Gurevych}]{tonglet2024image}
Jonathan Tonglet, Marie-Francine Moens, and Iryna Gurevych. 2024.
\newblock "image, tell me your story!" predicting the original meta-context of visual misinformation.
\newblock \emph{arXiv preprint arXiv:2408.09939}.

\bibitem[{Wang et~al.(2024)Wang, Wang, Li, Guan, and Li}]{wang2024harmfully}
Bing Wang, Shengsheng Wang, Changchun Li, Renchu Guan, and Ximing Li. 2024.
\newblock Harmfully manipulated images matter in multimodal misinformation detection.
\newblock In \emph{Proceedings of the 32nd ACM International Conference on Multimedia}, pages 2262--2271.

\bibitem[{Wang(2017)}]{wang2017liar}
William~Yang Wang. 2017.
\newblock " liar, liar pants on fire": A new benchmark dataset for fake news detection.
\newblock \emph{arXiv preprint arXiv:1705.00648}.

\bibitem[{Webhose.io(2024)}]{webhose}
Webhose.io. 2024.
\newblock \href {https://github.com/Webhose/free-news-datasets} {Free news datasets}.

\bibitem[{Weikmann and Lecheler(2023)}]{weikmann2023visual}
Teresa Weikmann and Sophie Lecheler. 2023.
\newblock Visual disinformation in a digital age: A literature synthesis and research agenda.
\newblock \emph{New Media \& Society}, 25(12):3696--3713.

\bibitem[{Wolf(2020)}]{wolf2020transformers}
Thomas Wolf. 2020.
\newblock Transformers: State-of-the-art natural language processing.
\newblock \emph{arXiv preprint arXiv:1910.03771}.

\bibitem[{Yoon et~al.(2024)Yoon, Yoon, and Park}]{yoon2024understanding}
Yejun Yoon, Seunghyun Yoon, and Kunwoo Park. 2024.
\newblock Understanding news thumbnail representativeness by counterfactual text-guided contrastive language-image pretraining.
\newblock \emph{arXiv preprint arXiv:2402.11159}.

\bibitem[{Zhang et~al.(2025)Zhang, Li, Chen, Ge, Sun, Zhang, Jiang, Yuan, Peng, and Luo}]{zhang2025flashvideo}
Shilong Zhang, Wenbo Li, Shoufa Chen, Chongjian Ge, Peize Sun, Yida Zhang, Yi~Jiang, Zehuan Yuan, Binyue Peng, and Ping Luo. 2025.
\newblock Flashvideo: Flowing fidelity to detail for efficient high-resolution video generation.
\newblock \emph{arXiv preprint arXiv:2502.05179}.

\end{thebibliography}

\appendix

~
\newpage

\section{Annotator Instructions}
\label{app:annotator-instructions}

These annotator instructions were posted both in the Prolific participant sourcing interface and in the Argilla annotation tool. The participants reviewed these instructions during paid response time.

\begin{tcolorbox}[colback=lightgray, colframe=black, title=Annotator Instructions]

This study involves reading short news articles and answering questions about the main images featured in these articles. The questions will ask you to identify the time and location of capture, based on the context provided in the article. The collected dataset will be open-sourced for use in ethical AI training.

Thank you for participating in our study! You will be presented with short news articles and asked to provide information about the images used in these articles. Specifically, for each image, you are asked to identify the most likely time and location of capture based on the article’s context and image caption. \\

{\small

\textbf{Time of Origin}

\begin{itemize}
    \item Provide the month and year when the image was most likely taken (e.g., “February 2024”, “November 2010”).
    \item If the month cannot be inferred, provide only the year (e.g., “2024”, “2010”).
    \item If the year cannot be inferred, respond with “N/A”.
\end{itemize}

\textbf{Location of Origin}

\begin{itemize}
    \item Provide the city and country where the image was most likely taken (e.g., “Boston, USA”, “Paris, France”).
    \item If the city cannot be inferred, provide only the country (e.g., “USA”, “France”).
    \item If the location cannot be determined, respond with “N/A”.
Your responses should be based on the context of the article. If you cannot safely infer the time or location, please use “N/A”.
\end{itemize}

}

Annotate all $55$ articles.

\end{tcolorbox}

\vfill

\section{Prompts}
\label{app:prompts}

\begin{tcolorbox}[colback=lightgreen, colframe=black, title=Alternative Metadata Generation \\ (System Prompt)]
    You are a generator of places and locations that are absolutely unrelated to those presented.
\end{tcolorbox}

\begin{tcolorbox}[colback=lightgreen, colframe=black, title=Alternative Metadata Generation \\ (Inference Prompt)]
    Give me a place and a time that are absolutely unrelated to the following: \texttt{'\{ORIGINAL PLACE\}; \{ORIGINAL TIME\}'}. 
    
    Give your response in the same format: \texttt{'\{NEW PLACE\}; \{NEW TIME\}'}, and don't say anything else.
\end{tcolorbox}

\begin{tcolorbox}[colback=lightgreen, colframe=black, title=Benchmarking (System Prompt)]

You are evaluating the relevance and credibility of images and videos attached to news stories.

Below, you will be presented with:

\begin{itemize}
    \item The title and the body of the article
    \item The image caption, as presented in the article
    \item Provenance metadata indicating source location and time of the image
\end{itemize}

\end{tcolorbox}

~

\vfill

~

\begin{tcolorbox}[colback=lightgreen, colframe=black, title=Benchmarking (Inference Prompt)]
Here is the data:

\begin{itemize}
    \item The title: \texttt{TITLE}
    \item The body: \texttt{BODY}
    \item Image caption: \texttt{IMAGE CAPTION}
    \item (Provenance metadata) Image location: \texttt{SOURCE LOCATION}
    \item (Provenance metadata) Image time: \texttt{SOURCE TIME}
\end{itemize}

Analyze the following:

\begin{enumerate}
    \item Is the location where the image was taken relevant to the news story? Return yes or no.
    \item Is the time (year and month) when the image was taken relevant to the news story? Return yes or no.
\end{enumerate}

Respond in the following comma-separated format: \texttt{\{yes/no\}}, \texttt{\{yes/no\}}. Possible responses include: \texttt{'yes,yes'},  \texttt{'no,no'},  \texttt{'yes,no'}, or \texttt{'no,yes'}. Do not enumerate these or add any extra characters.
    
\end{tcolorbox}

\newpage

\section{Qualitative Results: Article Example}
\label{app:qual-res-article}

The following is an excerpt of the article used in the qualitative evaluation (Section~\ref{sec:results-qual}). It was published on June 13, 2024, on \url{www.nbcnewyork.com}. We include this excerpt under fair use to demonstrate the reasoning abilities of evaluated LLMs on LOR and DTOR.

\noindent\rule{\linewidth}{0.4pt}

\begin{figure}[h]
    \centering
    \includegraphics[width=\linewidth]{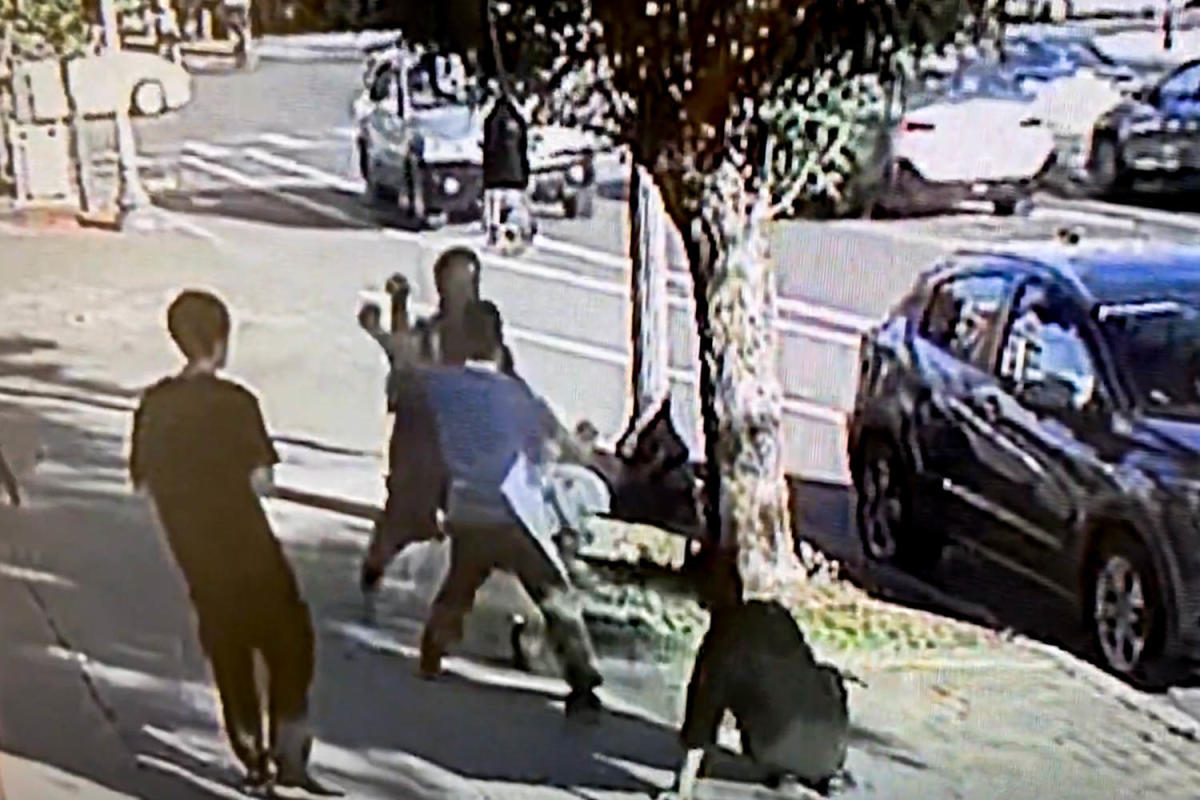}
\end{figure}

\noindent \textbf{Title:} N.Y. sushi restaurant owner out of coma after dine-and-dash attack over \$425 bill

\noindent \textbf{Body:} An Albany sushi restaurant owner is slowly showing signs of recovery after a brutal attack outside his restaurant last month. Su Wen, owner and chef at Shogun Sushi in upstate New York, has woken up from a nearly two-week coma and is experiencing increasing periods of consciousness, said Ray Ren, one of the managers at his restaurant...

\noindent\rule{\linewidth}{0.4pt}

\noindent \textbf{Provenance Metadata:}

\textit{Location of Origin:} Albany, USA

\textit{Date of Origin:} May, 2024

\onecolumn
\newpage
\captionsetup{justification=raggedright,singlelinecheck=false}

\section{Annotation Tool Screenshots}
\label{app:annotation-tool-screenshots}

\begin{figure}[H]
    
    \includegraphics[width=0.75\linewidth]{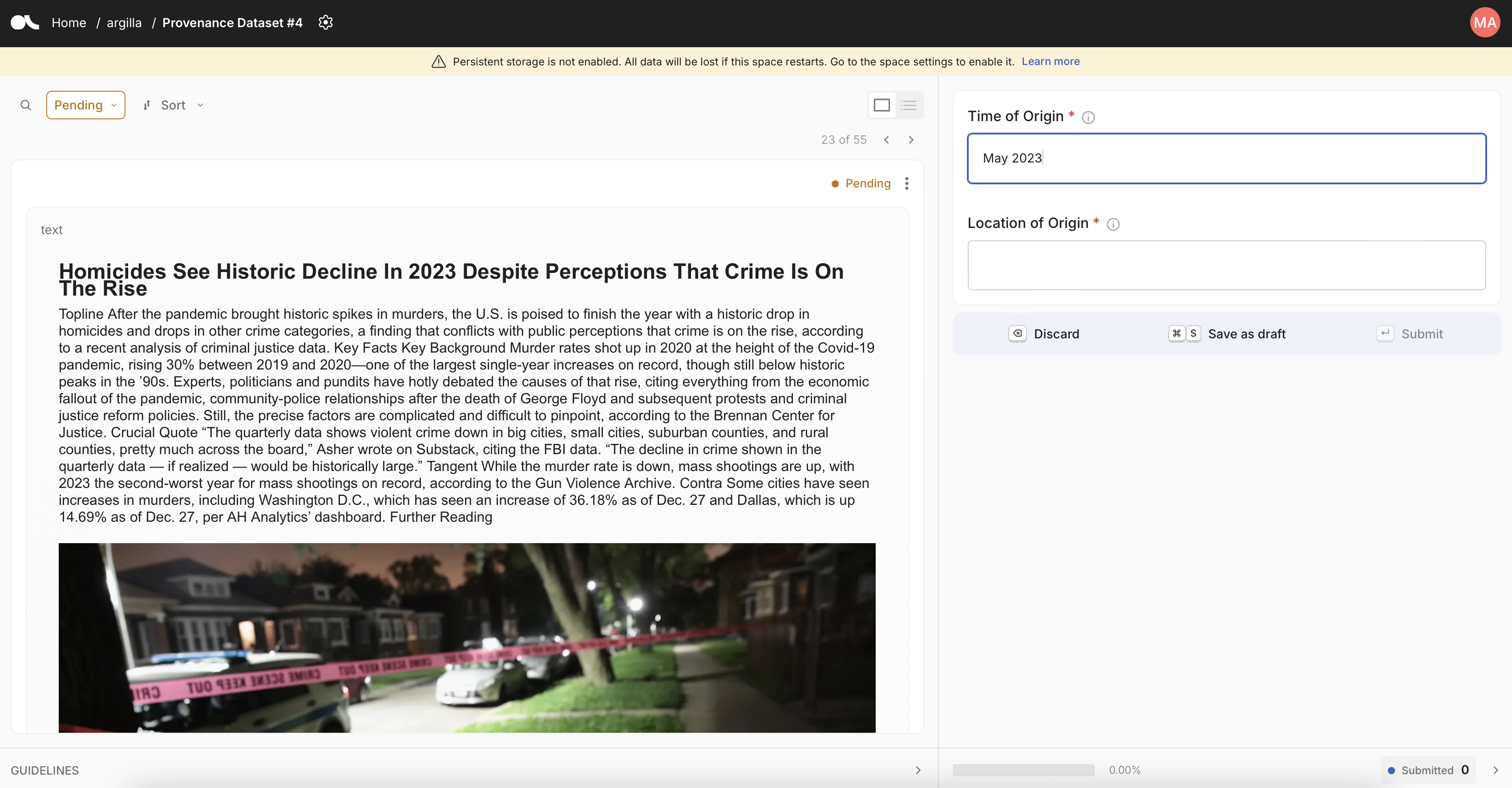}
    \caption{Screenshot of the Argilla annotation tool, focused on an article body.}
    \label{fig:ann-tool-0}
\end{figure}

\begin{figure}[H]
    
    \includegraphics[width=0.75\linewidth]{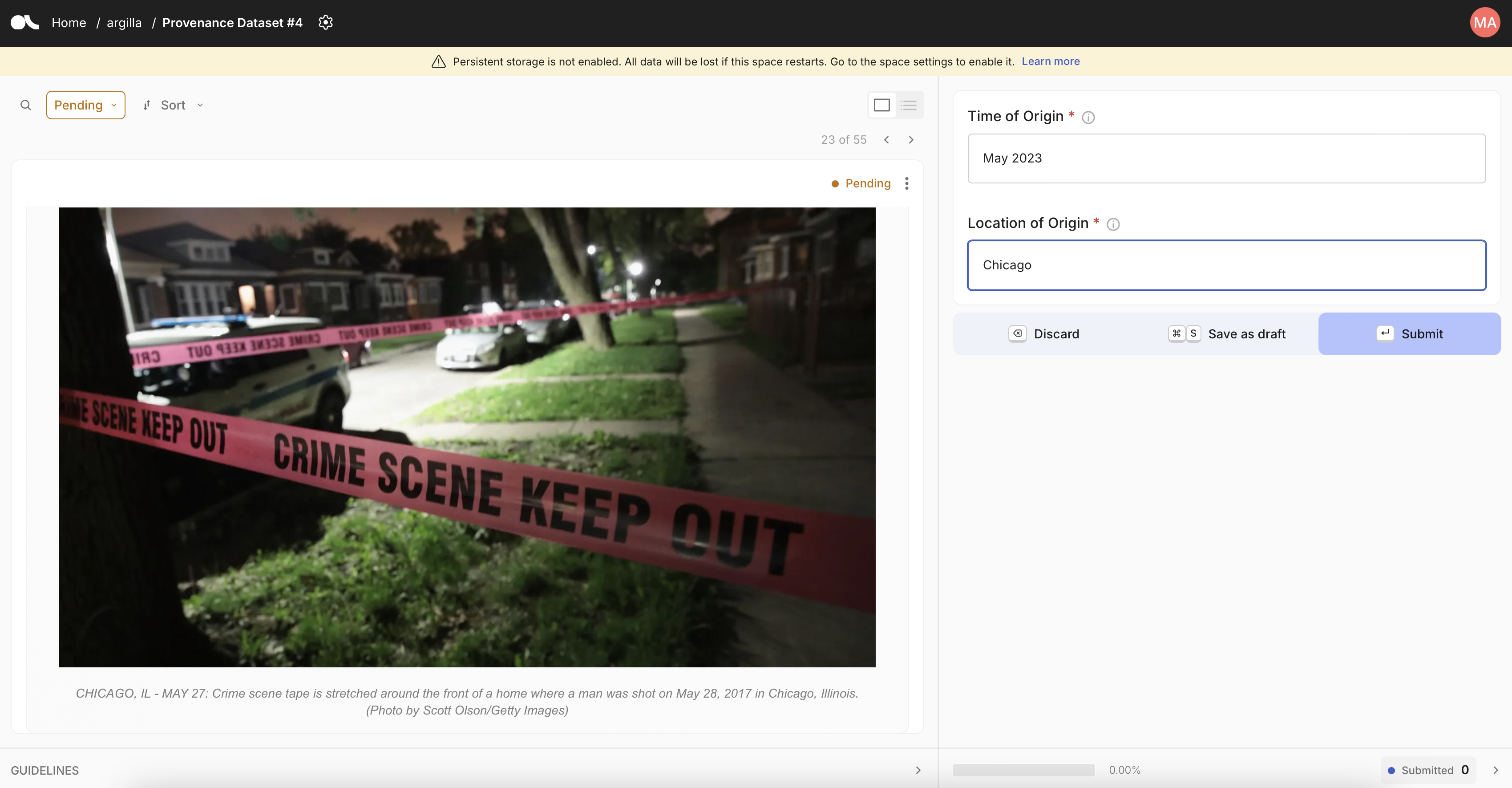}
    \caption{Screenshot of the Argilla annotation tool, focused on an image and its caption.}
    \label{fig:ann-tool-1}
\end{figure}

\begin{figure}[H]
    
    \includegraphics[width=0.75\linewidth]{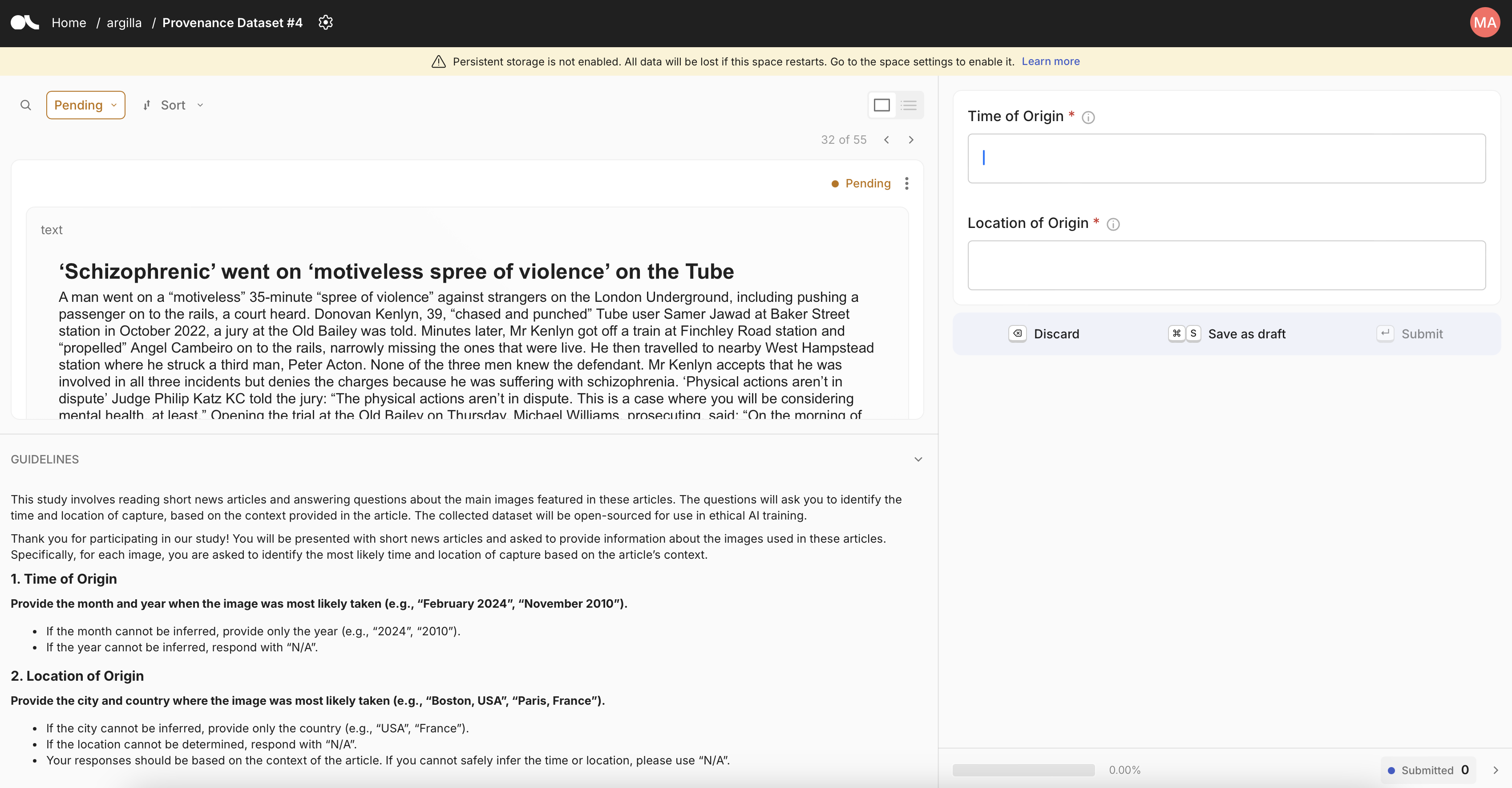}
    \caption{Screenshot of the Argilla annotation tool with the instructions window open.}
    \label{fig:ann-tool-2}
\end{figure}

\end{document}